\documentclass[runningheads]{llncs}
\usepackage{graphicx}

\usepackage{adjustbox}

\usepackage{subcaption}
\usepackage{hhline}

\begin{document}
\title{SqueezeFacePoseNet: \\ Lightweight Face Verification Across Different Poses for Mobile Platforms}
%
%
\author{Fernando Alonso-Fernandez\inst{1} \and
Javier Barrachina\inst{2} \and
Kevin Hernandez-Diaz\inst{1} \and
Josef Bigun\inst{1}}
\authorrunning{F. Alonso-Fernandez et al.}
%
\institute{School of Information Technology, Halmstad University, Sweden \\ \email{feralo@hh.se, kevher@hh.se, josef.bigun@hh.se} \and
Facephi Biometria, Av. México 20, Edificio Marsamar, 03008 Alicante, Spain \\
\email{jbarrachina@facephi.com}}
\maketitle              
\begin{abstract}
Ubiquitous and real-time person authentication has become critical after the breakthrough of all kind of services provided via mobile devices.
In this context, face technologies can provide reliable and robust user authentication, given the availability of cameras in these devices, as well as their
widespread use in everyday applications.
The rapid development of deep Convolutional Neural Networks (CNNs) has resulted in many accurate face verification architectures. However, their typical size (hundreds of megabytes) makes them infeasible to be incorporated in downloadable mobile applications where the entire file typically may not exceed 100 Mb.
Accordingly, we address the challenge of developing a lightweight face recognition network of just a few megabytes that can operate with sufficient accuracy in comparison to much larger models.
The network also should be able to operate under different poses, given the variability naturally observed in uncontrolled environments where mobile devices are typically used.
In this paper, we adapt the lightweight SqueezeNet model, of just 4.4MB, to effectively provide cross-pose face recognition.
After trained on the MS-Celeb-1M and VGGFace2 databases, our model achieves an EER of 1.23\% on the difficult frontal vs. profile comparison, and 0.54\% on profile vs. profile images.
Under less extreme variations involving frontal images in any of the enrolment/query images pair, EER is pushed down to $<$0.3\%, and the FRR at FAR=0.1\% to less than 1\%.
This makes our light model suitable for face recognition where at least acquisition of the enrolment image can be controlled.
At the cost of a slight degradation in performance, we also test an even lighter model (of just 2.5MB) where regular convolutions are replaced with depth-wise separable convolutions.

\end{abstract}

\section{Introduction}

All kind of services are migrating from physical to digital domains.
Mobiles have become data hubs, storing sensitive data like
payment information, photos, emails or passwords \cite{[Akhtar18-QAbiometrics]}.
In this context,
biometric technologies hold a great promise to provide
reliable and robust user authentication using the sensors embedded in such devices \cite{[Jain16]}.
But in order for algorithms to operate with sufficient accuracy,
they need to be adapted to the limited processing resources of mobile devices.
Data templates also have to be small if they are to be transmitted.
On top of it, mobile environments usually imply little control in the acquisition
(e.g. on-the-move or on-the-go), leading to huge variability in data quality.

In this work, we are interested in face technologies in mobile environments.
Face verification is increasingly used in applications such as
device unlock, mobile payments, login to applications, etc.
Recent developments involve deep learning \cite{[Sundararajan18-DLbiometrics]}.
Given enough data, they generate classifiers with impressive performance
in unconstrained scenarios with high variability.
However, state-of-the-art solutions are built upon
big deep Convolutional Neural Networks (CNNs), e.g. \cite{[He16]},
with dozens of millions of parameters and models that typically occupy hundreds of megabytes.
Such a big size and the computational resources that such networks require make them unfeasible for embedded mobile applications.

In recent years, lighter CNN architectures have been proposed for common visual tasks,
e.g. MobileNet \cite{[Howard17MobileNetv1corr]}, MobileNetV2 \cite{[Sandler18mobilenetv2]}, ShuffleNet \cite{[Zhang18ShuffleNet]} or SqueezeNet \cite{[Iandola16SqueezeNet]}.
%
%
Several works have bench-marked some of these networks for face recognition \cite{[Chen18MobileFaceNets],[Duong19MobiFace],[Martinez19ShuffleFaceNet]}.
Even if they employ training databases that contain images captured under a wide range of variations,
these works have not specifically assessed 
performance across different poses.
In this work, our main contribution is therefore a novel lightweight face recognition network
which is tested against a database specifically designed to explore pose variations \cite{[Cao18vggface2]}.
We base our developments on SqueezeNet, which is a much lighter architecture than the other networks.
To the best of our knowledge, this is the first work testing deep face recognition performance specifically under different poses and in mobile environments.
With a database of 11040 images from 368 subjects, 
our experiments show that the proposed network compares well against two larger benchmark networks having a size $>$30 times bigger and $>$20 times more parameters.

\begin{table*}[htb]
\begin{center}
\begin{tabular}{|c|c|c|c|c|c|c|}

\multicolumn{7}{c}{} \\ \hline

& Input & & Model & Para- & Vector &   Inference \\

Network & size & Layers & Size & meters & Size &   Time \\

\hline

\multicolumn{7}{c}{} \\

\multicolumn{7}{c}{Existing lightweight CNN architectures for face recognition} \\ \hline

LightCNN \cite{[Wu18lightCNN]} & 128$\times$128 & 29 & n/a & 12.6M & 256 &  n/a \\ \hline

MobileFaceNets \cite{[Chen18MobileFaceNets]} & 112$\times$112 & 50 & 4MB & 0.99M & 256 & 24ms (*) \\ \hline

MobiFace \cite{[Duong19MobiFace]} & 112$\times$112 & 45 & 11.3MB & n/a & 512 & 28ms (*) \\ \hline

ShuffleFaceNet \cite{[Martinez19ShuffleFaceNet]} & 112$\times$112 & n/a  & 10.5MB & 2.6M & 128  &   29.1ms (*) \\ \hline

SeesawFaceNets \cite{[zhang19seesawfacenets]} & 112$\times$112 & 50 & n/a & 1.3M & 512 &  n/a \\ \hline

\multicolumn{7}{c}{} \\

\multicolumn{7}{c}{Networks evaluated in the present paper} \\ \hline

\textbf{SqueezeFacePoseNet} & 113$\times$113 & 18 & 4.41MB & 1.24M & 1000 &  37.7ms \\ \hline

\multicolumn{1}{|r|}{\textbf{+GDC}} & 113$\times$113 & 18 & 5.01MB & 1.4M & 1000 &   38.7ms \\ \hline

\multicolumn{1}{|r|}{\textbf{+DWC}} & 113$\times$113 & 18 & 2.5MB & 0.69M & 1000 &   36.4ms \\ \hline

\multicolumn{1}{|r|}{\textbf{+DWC+GDC}} & 113$\times$113 & 18 & 3.1MB & 0.86M & 1000 &  36.9ms  \\ \hhline{=======}

ResNet50ft \cite{[Cao18vggface2]} & 224$\times$224 & 50 & 146MB & 25.6M  & 2048 & 0.16s \\ \hline

SENet50ft \cite{[Cao18vggface2]} & 224$\times$224 & 50 & 155MB & 28.1M  & 2048 & 0.21s \\ \hline

\end{tabular}

\end{center}
\caption{Top: proposed lightweight models in the literature for face recognition. Bottom: networks evaluated in the present paper. (*) Inference times are as reported in the respective papers, so they are not fully comparable. The hardware used in the reported studies includes
a Qualcomm Snapdragon 820 mobile CPU @ 2.2 GHz \cite{[Chen18MobileFaceNets]},
an Intel i7-6850K CPU @ 3.6GHz \cite{[Duong19MobiFace]},
and an Intel i7-7700HQ CPU @ 2.80 GHz \cite{[Martinez19ShuffleFaceNet]}.
%
The latter also carries out a comparison of different devices, including high-end GPUs, with inference times reduced around one order of magnitude.
%
Please refer to the original papers for details. Inference in this paper is done with
an Intel i7-8650U CPU @ 1.9GHz. 
}
\label{tab:networks}
\end{table*}
\normalsize

\section{Related Works}

Lightweight CNNs
%
%
employ different techniques to achieve less parameters and faster processing, such as point-wise convolution, depth-wise separable convolution, 
and bottleneck layers.
Point-wise convolutions consist of 1$\times$1 filters with a depth equal to the number of input channels, and it is used to reduce or augment the number of channels.
Depth-wise separable convolution splits convolution in two steps,
the first one performing lightweight filtering by using a single convolutional filter per input channel, followed by a 1$\times$1 point-wise convolution that carries out linear combinations of the input channels.
For single convolutional filters of 3 $\times$ 3, depth-wise separable convolution achieves a computational reduction of 8-9 times in comparison to standard convolution, with a small cost in accuracy only \cite{[Howard17MobileNetv1corr]}.
Bottleneck layers consist on obtaining a representation of the input with reduced dimensionality before processing it with a larger amount of filters that usually have bigger spatial dimensions as well.

SqueezeNet is one of the early works presenting an architecture with fewer parameters and a smaller size (1.24M parameters, 4.6MB, and 18 convolutional layers). The authors proposed 1$\times$1 point-wise convolutions with squeeze and expand modules that follow the bottleneck concept.
Later, MobileNet (4.M parameters) and MobileNetV2 (3.5M parameters, 13MB, and 53 convolutional layers) were proposed. The former uses faster depth-wise and point-wise convolutions,
and the latter uses bottlenecks and inverted residual structures.
Inverted residual structures consist of adding a shortcut between bottleneck layers, similar to residual connections \cite{[He16]}, that allows to reuse features through the network and to improve the ability of a gradient to propagate across multiple layers.
Lastly, ShuffleNet (1.4M parameters, 6.3MB, and 50 convolutional layers) employs point-wise group convolution and channel shuffle to reduce the computational cost.
%

Some works have designed light face recognition models based on these or other architectures (Table~\ref{tab:networks}).
To carry out biometric verification, they typically use as feature vector the output before the fully-connected part.
The authors in \cite{[Wu18lightCNN]} presented LightCNN, with 29 convolutional layers and residual connections, which has 12.6M parameters.
With a compact vector of 256 elements, they achieved 99.33\% verification accuracy on the LFW database.
MobileFaceNets \cite{[Chen18MobileFaceNets]} is based on MobileNetV2 but with smaller expansion factors on bottleneck layers, obtaining a network of 0.99M parameters and 4MB. The authors introduced Global Depth-wise Convolution (GDC) to substitute the standard Global Average Pooling (GAP) at the end of the network. The motivation is that GAP treats all pixels of the last channels equally, but in face recognition, the center pixels should not have the same role than corner pixels.
They also used PReLU as non-linearity, and fast down-sampling at the beginning of the network.
With a vector of 256 element, the reported accuracy on LFW was 99.55\%.
MobiFace \cite{[Duong19MobiFace]} is also based on MobileNetV2. Besides fast down-sampling and PReLU, they change GAP by a fully-connected layer in the last stage of the embedding to allow learning of different weights for each spatial region of the last channels.
With a network of 11.3MB and a vector of 512 elements, the reported accuracy on LFW was 99.73\%.
ShuffleNet is used as base for ShuffleFaceNet \cite{[Martinez19ShuffleFaceNet]}.
Here the authors also use PReLU, and replace GAP with GDC.
They test a different number of channels in each block, and the network with the best speed-accuracy trade-off has a size of 10.5MB and 2.6M parameters, with a feature model of 128 elements.
The reported accuracy on LFW is of 99.67\%.
Lastly, the work \cite{[zhang19seesawfacenets]} presented SeesawFaceNets, based on seesaw blocks \cite{[Zhang09SeesawNet]}.
Based on inverted residual bottleneck blocks, seesaw blocks replace
point-wise convolutions with uneven group convolutions
and channel permute/shuffle operations.
The author also added Squeeze-and-Excitation (SE) \cite{[Hu18]}, and used Swish as non-linearity.
SE blocks explicitly model channel relationships in order to
adaptively recalibrate channel-wise feature responses,
and they can be integrated with many architectures, improving their representation power.
With a network of 1.3M parameters
and vectors of 512 elements,
the author reported an accuracy on LFW of 99.7\%.

\section{Network Architecture}

\begin{figure}[htb]
\centering
        \includegraphics[width=0.45\textwidth]{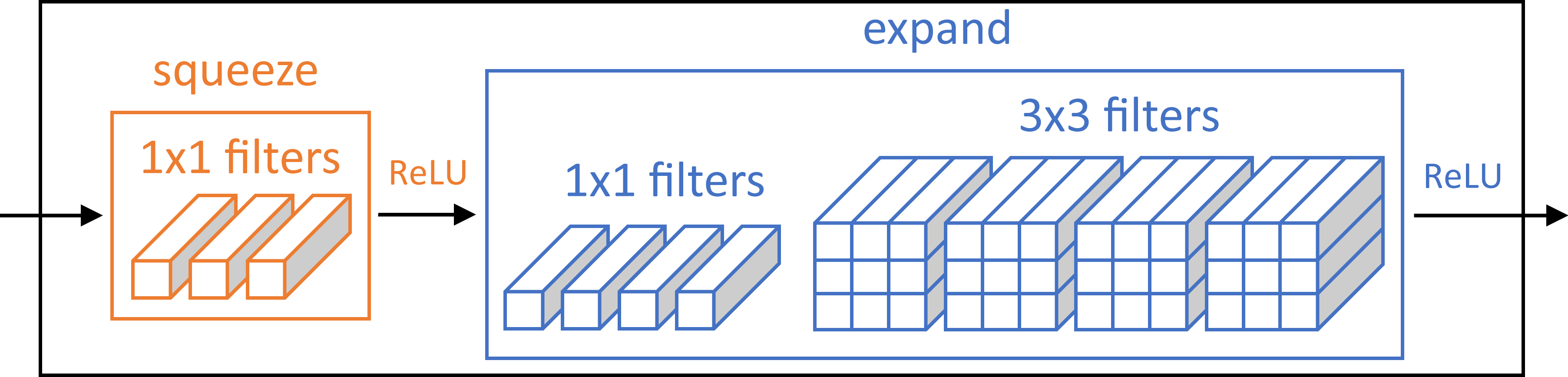}
\caption{Internal architecture of a fire module. In this example, the squeeze layer has three 1$\times$1 filters, and the expand layer has four 1$\times$1 and four 3$\times$3 filters. Adapted from \cite{[Iandola16SqueezeNet]}.}
\label{fig:fire_layer}
\end{figure}

As back-bone model, we employ SqueezeNet \cite{[Iandola16SqueezeNet]}.
This is the smallest architecture among the generic light CNNs mentioned.
With only 1.24M parameters and 4.6 MB in its uncompressed version, it matched AlexNet accuracy on ImageNet with 50x fewer parameters.
%
%
Its building brick, 
called fire module (Figure~\ref{fig:fire_layer}), contains two layers: a squeeze layer and an expand layer.
The squeeze layer uses 1$\times$1 (point-wise) filters as a bottleneck to reduce dimensionality of the feature maps that will be processed in the expand layer with (more costly) 3$\times$3 filters.
%
Also, to achieve further parameter reduction, some filters in the expand layer are of 1$\times$1 instead of 3$\times$3. 
The squeezing (bottleneck) and expansion behavior is common in CNNs, helping to reduce the amount of parameters while keeping the same feature map size between the input and output \cite{[Sandler18mobilenetv2]}.
In addition, SqueezeNet uses late downsampling, so many convolution layers have large activation
maps. Intuitively, this should lead to a higher accuracy.
The architecture of the employed network is shown in Table~\ref{tab:squeezenet-architecture}, which mirrors \cite{[Iandola16SqueezeNet]} with slight changes.
%
%

The network has been modified to employ an input size of 113$\times$113$\times$3.
It starts with a convolutional layer with 64 filters of 3$\times$3$\times$3 (the original paper uses 96 filters), followed by 8 fire modules.
The stride of the first convolutional layer has been changed from 2 to 1, so the rest of the network can remain unchanged.
Then, the network ends with a convolutional layer with 1000 filters of 1$\times$1$\times$512.
ReLU is applied after each convolutional layer, and dropout of 50\% is applied after the last fire module.
All convolutional layers have stride 1, and all max-pooling layers are of 3$\times$3 and stride 2.
As it can be observed, the number of filters in each fire module increases gradually.
Also, the network uses GAP, which carries out down-sampling by computing the average of each input channel.
This reduces the input size to the classification layer.
After GAP, we add a fully connected layer that matches as output size the number of classes of the training database. Batch-normalization and dropout at 50\% is also added to counteract over-fitting in the fully connected layer due to the high number of training classes (35K and 8.6K).
We will refer to this network as SqueezeFacePoseNet.
To achieve an even smaller model, we will test the replacement of standard convolution with depth-wise separable convolution in all 3$\times$3 filters, and we will also evaluate the replacement of GAP with GDC.
The size and amount of parameters of the different combinations is shown in Table~\ref{tab:networks}, bottom.

\begin{table}[htb]
\begin{center}
\begin{tabular}{ccccc}

\multicolumn{5}{c}{} \\

& output & $\#$1$\times$1  & $\#$1$\times$1  & $\#$3$\times$3  \\
layer & size & squeeze & expand & expand \\ \hline

input & 113$^2$$\times$3 & - & - & - \\ \hline
conv1 & 113$^2$$\times$64 & - & - & - \\ \hline
maxpool1 & 56$^2$$\times$64 & - & - & - \\ \hline
fire2 & 56$^2$$\times$128 & 16 & 64 & 64 \\ \hline
fire3 & 56$^2$$\times$128 & 16 & 64 & 64 \\ \hline
fire4 & 56$^2$$\times$256 & 32 & 128 & 128 \\ \hline
maxpool4 & 27$^2$$\times$256 & - & - & - \\ \hline
fire5 & 27$^2$$\times$256 & 32 & 128 & 128 \\ \hline
fire6 & 27$^2$$\times$384 & 48 & 192 & 192 \\ \hline
fire7 & 27$^2$$\times$384 & 48 & 192 & 192 \\ \hline
fire8 & 27$^2$$\times$512 & 64 & 256 & 256 \\ \hline
maxpool8 & 13$^2$$\times$512 & - & - & - \\ \hline
fire9 & 13$^2$$\times$512 & 64 & 256 & 256 \\ \hline
dropout9 & 13$^2$$\times$512 & - & - & - \\ \hline
conv10 & 13$^2$$\times$1000 & - & - & - \\ \hline
averagepool10 & 1$^2$$\times$1000 & - & - & - \\ \hline
batchnorm10 & 1$^2$$\times$1000 & - & - & - \\ \hline
dropout10 & 1$^2$$\times$1000 & - & - & - \\ \hline
fc & 1$^2$$\times$ C & - & - & - \\ \hline
softmax & 1$^2$$\times$ C & - & - & - \\ \hline

\end{tabular}

\end{center}
\caption{Architecture of the network used. C is the number of training classes.}
\label{tab:squeezenet-architecture}
\end{table}
\normalsize

%
%
%
%
%
%
%

We also evaluate the CNNs used in \cite{[Cao18vggface2]} to assess face recognition performance with the VGGFace2 database.
They use ResNet50 \cite{[He16]} and SE-ResNet50 \cite{[Hu18]} as backbone architectures, both with 50 convolutional layers, and ending with a GAP layer with produces a vector of 2048 elements before the fully connected layer.
ResNet networks presented the concept of residual connections to ease the training of CNNs.
%
%
%
The models employed in this paper\footnote{https://github.com/ox-vgg/vgg$\_$face2} are initialized from scratch,
then trained on the MS-Celeb-1M \cite{[Guo16_MSCeleb1M]} dataset,
and further fine-tuned on the VGGFace2 dataset.
We will refer to these as ResNet50ft and SENet50ft.

\begin{figure}[htb]
\centering
        \begin{subfigure}{.49\textwidth}
            \centering
            \includegraphics[width=.97\linewidth]{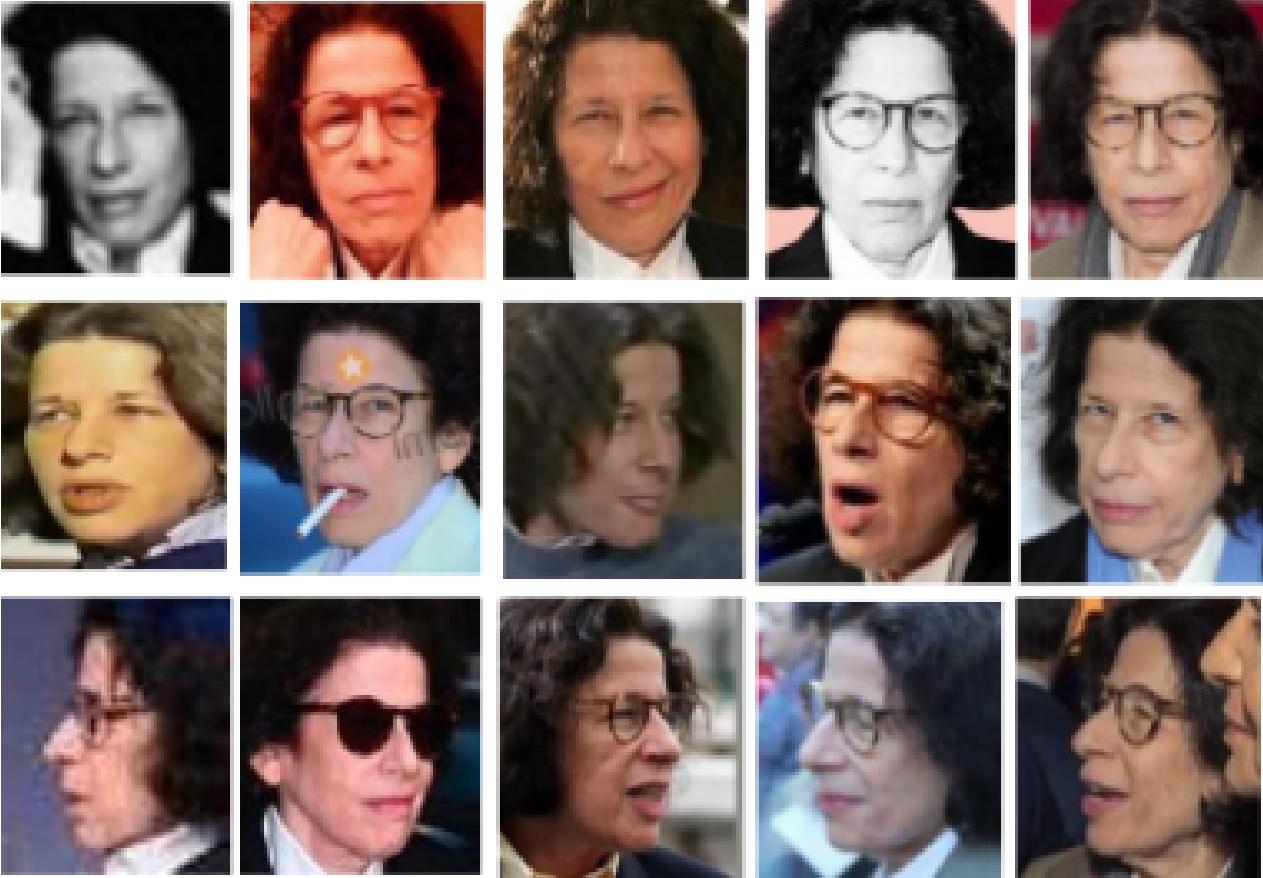}
            \caption{VGGFace2 pose templates from three viewpoints (frontal, three-quarter, and profile, arranged by row). Image from \cite{[Cao18vggface2]}.}
            \vspace{5mm}
        \end{subfigure}
        \hfill
        \begin{subfigure}{.49\textwidth}
            \centering
            \includegraphics[width=.97\linewidth]{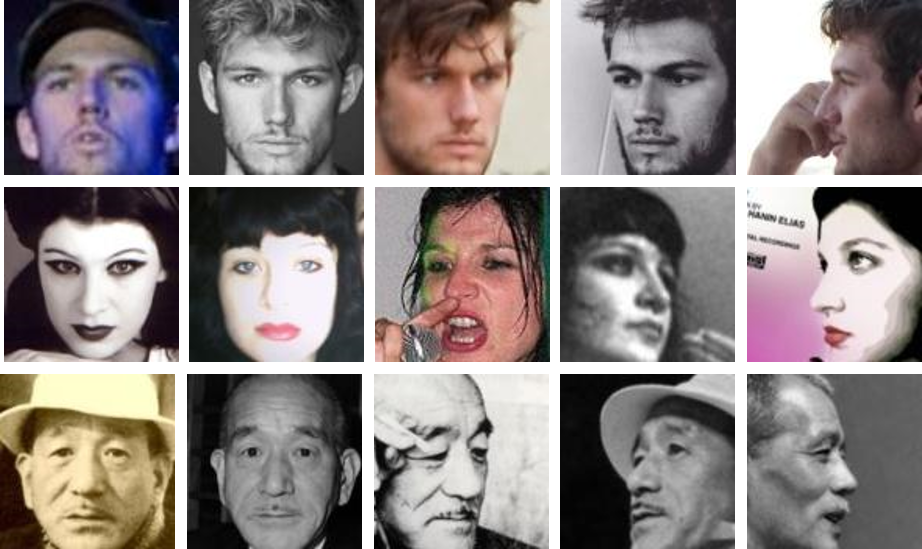}
            \caption{MS-Celeb-1M from three users (by row) and three profiles (by column: frontal (1-2), three-quarter (3-4), and profile (5)).}
            \vspace{2mm}
        \end{subfigure}
        \hfill
        \begin{subfigure}{.6\textwidth}
            \centering
            \includegraphics[width=.97\linewidth]{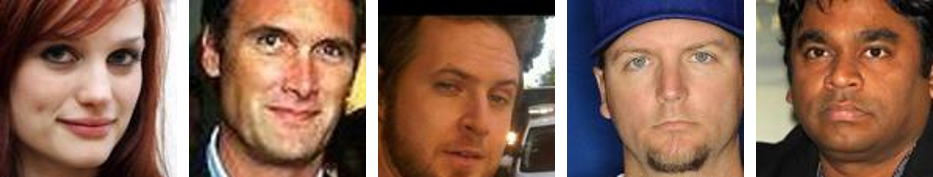}
            \caption{VGGFace2 training images with random crop.}
        \end{subfigure}
\caption{Example images of the databases employed.}
\label{fig:databases}
\end{figure}

\begin{table}[htb]
\centering

\begin{adjustbox}{max width=\textwidth}


\begin{tabular}{|c|c|c|c|c|c|c|}

\multicolumn{7}{c}{} \\ \cline{3-4} \cline{6-7}

\multicolumn{1}{c}{} & \multicolumn{1}{c}{} & \multicolumn{2}{|c|}{\textbf{SAME-POSE}}  & \multicolumn{1}{c}{} & \multicolumn{2}{|c|}{\textbf{CROSS-POSE}}  \\  \cline{3-4} \cline{6-7}

\multicolumn{1}{c}{template} &  \multicolumn{1}{c|}{} & genuine & impostor & \multicolumn{1}{c|}{} & genuine & impostor \\ \cline{1-1} \cline{3-4} \cline{6-7}

1 image & \multicolumn{1}{c|}{} &  368 $\times$ (9+8+...+1) = 16560  & 368 $\times$ 100 = 36800 & \multicolumn{1}{c|}{} & 368 $\times$ 10 $\times$ 10 = 36800 &  368 $\times$ 100 = 36800 \\ \cline{1-1} \cline{3-4} \cline{6-7}

5 images &  \multicolumn{1}{c|}{} & 368 $\times$ 1 = 368  & 368 $\times$ 100 = 36800 & \multicolumn{1}{c|}{} & 368 $\times$ 2 $\times$ 2 = 1472 &  368 $\times$ 100 = 36800 \\ \cline{1-1} \cline{3-4} \cline{6-7}

\multicolumn{7}{c}{} \\

\end{tabular}



\end{adjustbox}

\caption{Number of biometric verification scores.}
\label{tab:scores}

\end{table}


\section{Database and Experimental Protocol}

\label{subsec:db-protocol}

We use the VGGFace2 dataset, with 3.31M images of 9131 celebrities,
and an average of 363.6 images per person \cite{[Cao18vggface2]}.
The images, downloaded from the Internet, show large variations in pose, age, ethnicity,
lightning and background.
The database is divided into 8631 training classes (3.14M images), and the
remaining 500 for testing.
To enable recognition across different pose,
a subset of 368 subjects from the test set is provided (VGGFace2-Pose for short),
with 10 images per pose (frontal, three-quarter, and profile), totalling 11040 images.

To further improve recognition performance of our mobile network, we also use the RetinaFace cleaned set of the MS-Celeb-1M database \cite{[Guo16_MSCeleb1M]} to pre-train our model (MS1M for short).
Face images 
are pre-processed to a size of 112$\times$112 by five facial landmarks provided by RetinaFace \cite{[Deng19RetinaFace]}. In total, there are 5.1M images of 93.4K identities.
While MS1M has a larger number of images,
its intra-identity variation is limited due to an average of 81 images per person.
For this reason, we investigate the benefit of
first pre-training on a dataset with a large number of images (MS1M),
then fine-tune with more intra-class diversity (VGGFace2).
This is the protocol in \cite{[Cao18vggface2]},
and it has been shown to provide enhanced performance,
in comparison to training the models only with VGGFace2.
Some example images of these databases are shown in Figure \ref{fig:databases}.

Our network is trained for biometric identification using the soft-max function.
The network is initialized using ImageNet weights, since it has been shown that such transfer-learning strategy can provide equal or better performance than if initialized from scratch, while converging much faster \cite{[Kornblith19imagenet_transfer_better]}.
For training, the bounding box of VGGFace2 images are resized, so the shorter side has 256 pixels,
then a 224$\times$224 region is randomly cropped \cite{[Cao18vggface2]}.
To accommodate to the input size of the CNN, images of both databases are scaled to 113$\times$113.
SGDM is used as optimizer, with mini-batches of 128.
The initial learning rate is 0.01, which is decreased to 0.005, 0.001, and 0.0001
when the validation loss plateaus.
Also, the learning rate of newly added layers is multiplied by 10 during the epochs that the
global learning rate is 0.01.
Two percent of images of each user in the training set are set aside for validation.
To speed-up training and reduce parameters of the fully connected layer dedicated to
under-represented classes, we remove users from MS1M with less than 70 images,
resulting in 35016 users and 3.16M images.
This ensures also that at least one image per user is available in the validation set.
All experiments have been done in a stationary computer with an i9-9900 processor, 64 Gb RAM, and a NVIDIA RTX 2080 Ti GPU.
We carry out training using Matlab r2019a,
while the implementations of ResNet50ft and SENet50ft are run using MatConvNet.

We carry out verification experiments with the 368 subjects of VGGFace2-Pose.
To enable comparison with state-of-the-art, the test protocol follows the procedure of \cite{[Cao18vggface2]}.
A template is defined for each user, consisting of five faces with the same pose,
so two templates are available per user and per pose.
A template is represented by a single vector,
which is computed by averaging the descriptors given by the CNN of the faces in the template set.
To test the robustness of the employed networks,
we also carry out experiments using only one image as template.
During testing, VGGFace2 images are resized, so the shorter side has 256 pixels.
A 224$\times$224 crop of the center is then done (instead of a random crop),
followed by a resize to 113$\times$113.
To extract a face descriptor, the last layers of our network trained in identification mode are removed,
and the features are extracted from the GAP layer,
having dimensionality 1000.
A distance measure ($\chi^2$ in our case) is then used to obtain the similarity between two templates.
With ResNet50ft and SENet50ft architectures, we use as descriptor the output of the
layer adjacent to the classification layer, with dimensionality 2048.
Also, ResNet50ft and SENet50ft employ input images of 224$\times$224, so VGGFace2 images are kept in
this size when testing with these two networks.
%
%
%
%

\begin{figure}[htb]
\centering
\centering
\includegraphics[width=.8\textwidth]{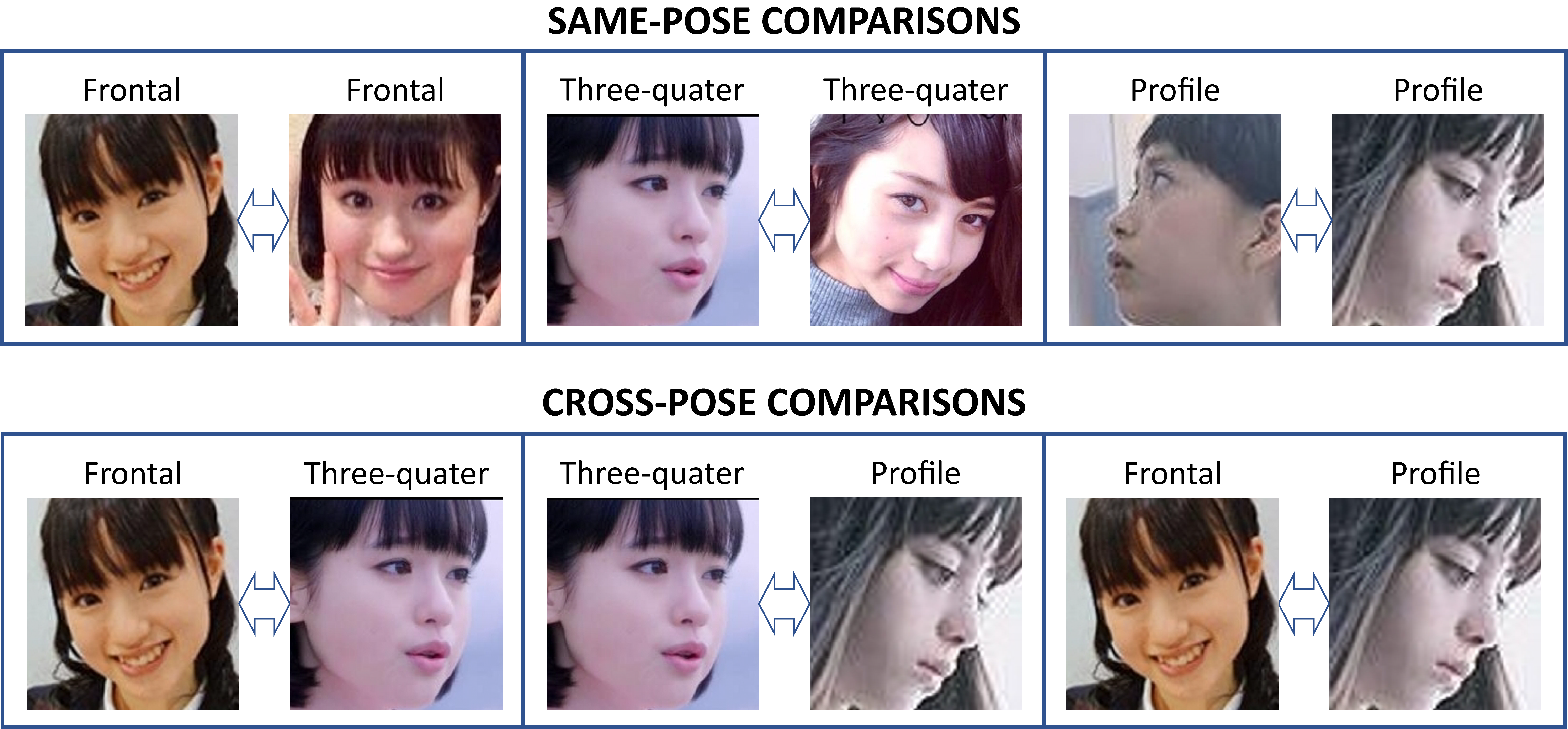}
\caption{Evaluation protocols: same-pose (left) and cross-pose comparisons (right).}
\label{fig:protocols}
\end{figure}

\begin{figure*}[htb]
\centering
    \includegraphics[width=0.98\textwidth]{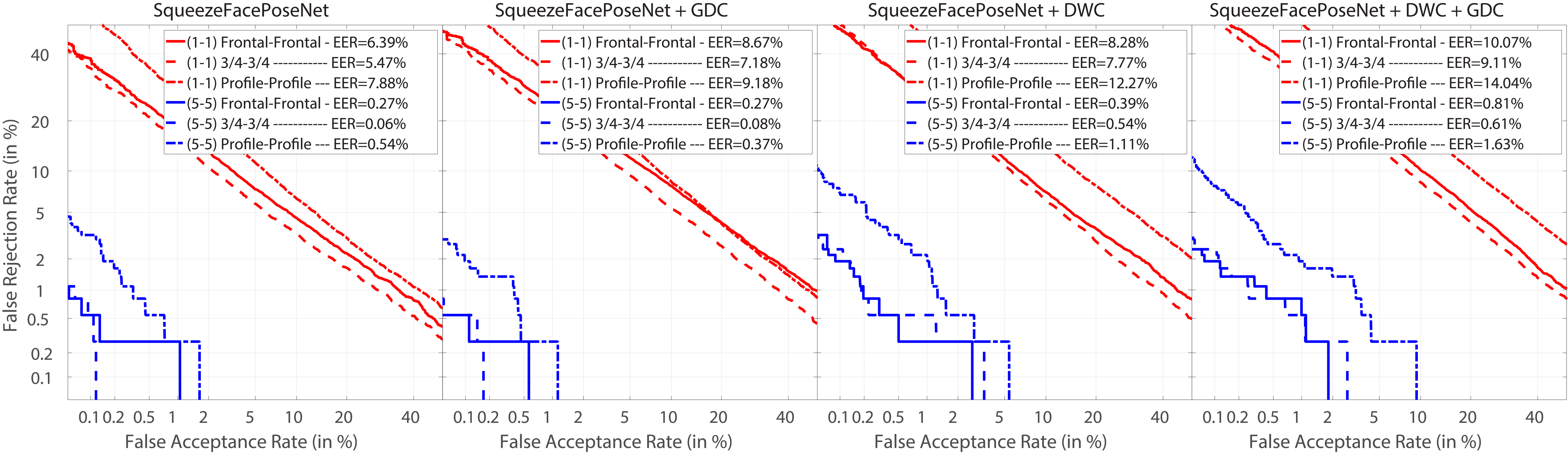}
\caption{SqueezeFacePoseNet: Face verification results (same-pose comparisons). Better in colour.}
\label{fig:same_pose_results_SqueezeFacePoseNet}
\end{figure*}

\begin{figure}[htb]
\centering
    \includegraphics[width=0.52\textwidth]{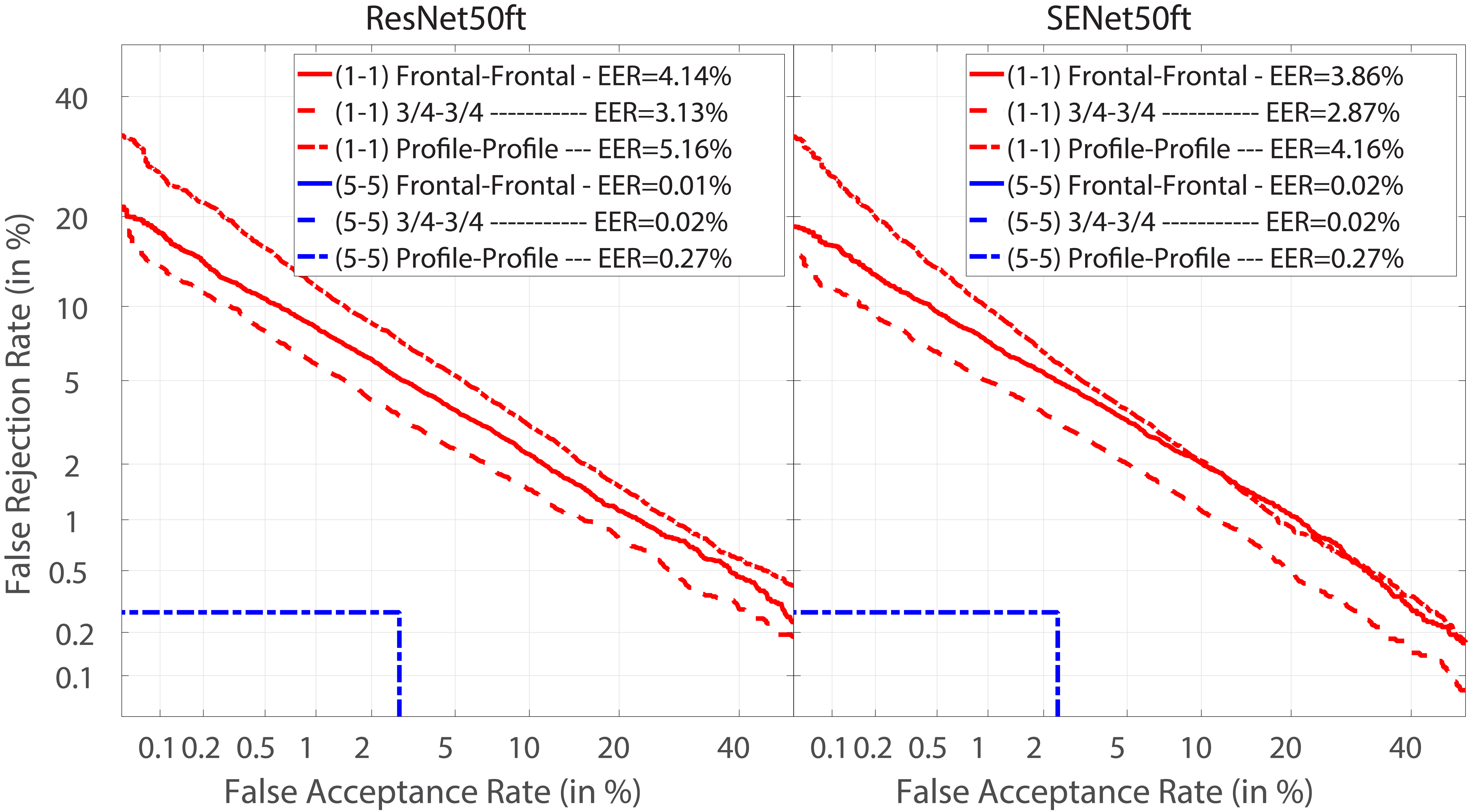}
\caption{ResNet50ft and SENet50ft (same-pose comparisons). Better in colour.}
\label{fig:same_pose_results_ResNet50ft_SENet50ft}
\end{figure}

\begin{figure*}[htb]
\centering
    \includegraphics[width=0.98\textwidth]{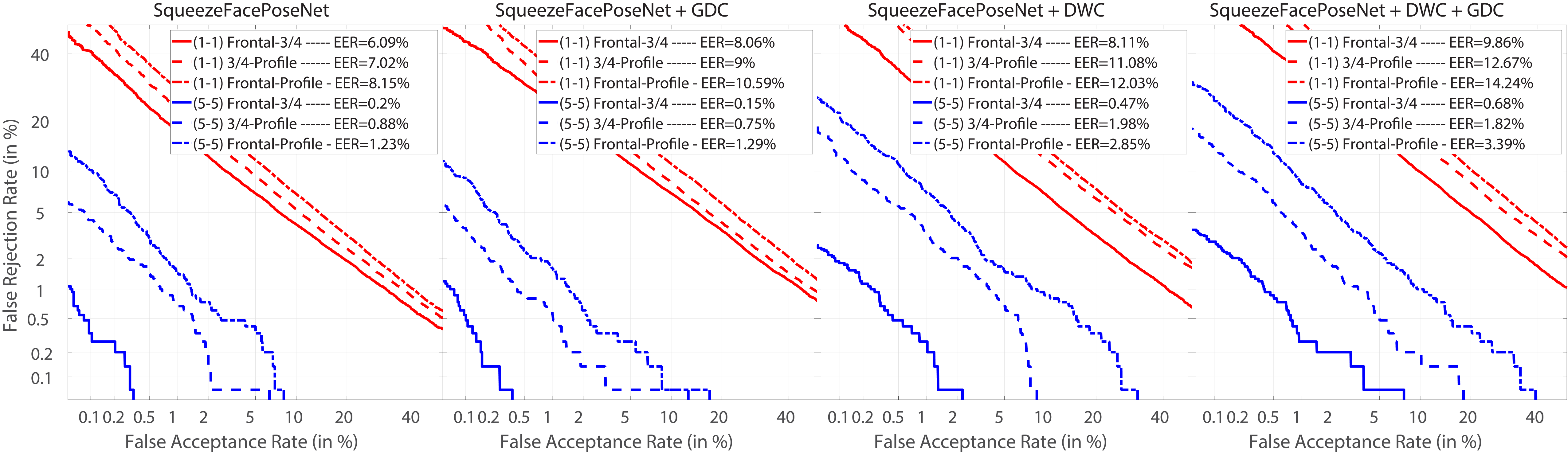}
\caption{SqueezeFacePoseNet: Face verification results (cross-pose comparisons). Better in colour.}
\label{fig:inter_pose_results_SqueezeFacePoseNet}
\end{figure*}

\begin{figure}[htb]
\centering
    \includegraphics[width=0.52\textwidth]{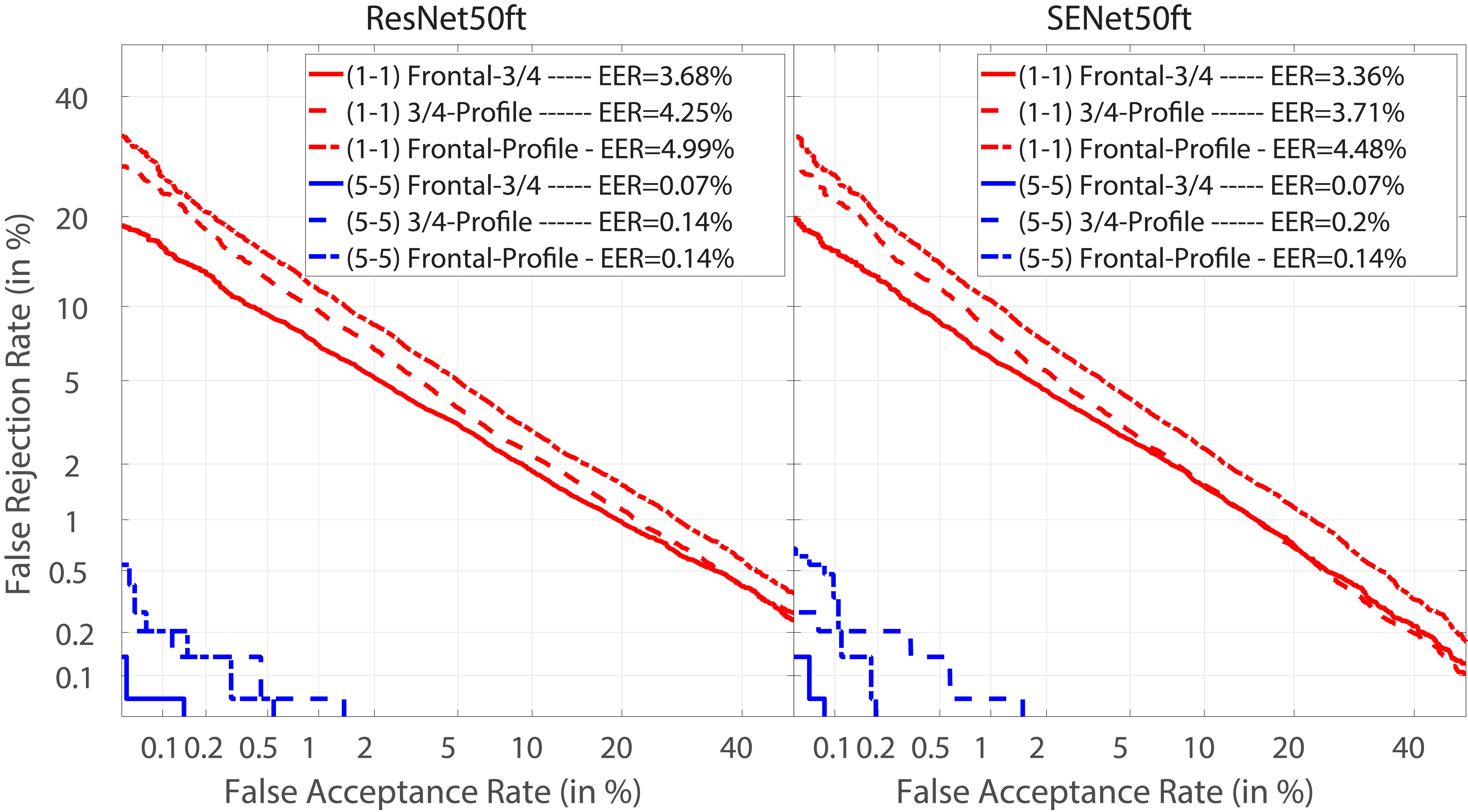}
\caption{ResNet50ft and SENet50ft (cross-pose comparisons). Better in colour.}
\label{fig:inter_pose_results_ResNet50ft_SENet50ft}
\end{figure}

\begin{table}[htb]
\centering

\begin{adjustbox}{max width=\textwidth}

\begin{tabular}{|c|c|ccc|ccc|c|ccc|ccc|}

\multicolumn{15}{c}{} \\ \cline{3-8} \cline{10-15}

\multicolumn{1}{c}{} &
\multicolumn{1}{c}{} &
\multicolumn{6}{|c|}{\textbf{One face image per template}} &
\multicolumn{1}{c}{} &
\multicolumn{6}{|c|}{\textbf{Five face images per template}}
\\ \cline{1-1} \cline{3-8} \cline{10-15}

\textbf{Recognition} &  & \multicolumn{3}{|c|}{\textbf{Same-Pose}} &  \multicolumn{3}{c|}{\textbf{Cross-Pose}} &  & \multicolumn{3}{|c|}{\textbf{Same-Pose}} &  \multicolumn{3}{c|}{\textbf{Cross-Pose}} \\  \cline{3-8} \cline{10-15}

\textbf{Network} &  & \textbf{F-F} & \textbf{3/4-3/4} & \textbf{P-P} &  \textbf{F-3/4} & \textbf{3/4-P} & \textbf{F-P} &  & \textbf{F-F} & \textbf{3/4-3/4} & \textbf{P-P} &  \textbf{F-3/4} & \textbf{3/4-P} & \textbf{F-P} \\ \hhline{=~======~======}

SqueezeFacePoseNet &  & 6.39 & 5.47 & 7.88 &  6.09 & 7.02 & 8.15 &  & 0.27 & 0.06 & 0.54 &  0.2 & 0.88 & 1.23 \\ \cline{1-1} \cline{3-8} \cline{10-15}

\multicolumn{1}{|r|}{+GDC} &  & 8.67 & 7.18 & 9.18 &  8.06 & 9 & 10.59 &  & 0.27 & 0.08 & 0.37 &  0.15 & 0.75 & 1.29 \\ \cline{1-1} \cline{3-8} \cline{10-15}

\multicolumn{1}{|r|}{+DWC} &  & 8.28 & 7.77 & 12.27 &  8.11 & 11.08 & 12.03 &  & 0.39 & 0.54 & 1.11 &  0.47 & 1.98 & 2.85 \\ \cline{1-1} \cline{3-8} \cline{10-15}

\multicolumn{1}{|r|}{+DWC+GDC} &  & 10.07 & 9.11 & 14.04 &  9.86 & 12.67 & 14.24 &  & 0.81 & 0.61 & 1.63 &  0.68 & 1.82 & 3.39 \\ \hhline{=~======~======}

ResNet50ft &  & 4.14 & 3.13 & 5.16 &  3.68 & 4.25 & 4.99 &  & \textbf{0.01} & \textbf{0.02} & 0.27 &  \textbf{0.07} & \textbf{0.14} & \textbf{0.14} \\ \cline{1-1} \cline{3-8} \cline{10-15}

SENet50ft &  & \textbf{3.86} & \textbf{2.87} & \textbf{4.16} &  \textbf{3.36} & \textbf{3.71} & \textbf{4.48} &  & 0.02 & \textbf{0.02} & \textbf{0.27} &  \textbf{0.07} & 0.2 & \textbf{0.14} \\ \cline{1-1} \cline{3-8} \cline{10-15}

\multicolumn{15}{c}{} \\

\end{tabular}


\end{adjustbox}

\caption{Face verification results on the VGGFace2-Pose database (EER \%). F=Frontal View. 3/4= Three-Quarter. P=Profile. The  best result of each column is marked in bold.}
\label{tab:results}

\end{table}

\section{Results}

%
%
%

\subsection{Same-Pose Comparisons}
\label{subsect:samepose}

We first report experiments of same-pose comparisons, i.e. comparing only templates generated with images having the same pose (Figure~\ref{fig:protocols}, left).
Genuine trials are done by comparing each template
of a user to the remaining templates of the same user,
avoiding symmetric comparisons.
Concerning impostor experiments,
the first template of a user is used as enrolment template, and compared with the second template of the next 100 users.
Table~\ref{tab:scores} (left) shows the total number of scores with this protocol.
Recall than when templates are generated using 5 images, there are only two templates available per user and per pose.
On the other hand, when templates are generated with only one image, there are ten templates per user and per pose.
Face verification results following this protocol are given in Figures~\ref{fig:same_pose_results_SqueezeFacePoseNet} and \ref{fig:same_pose_results_ResNet50ft_SENet50ft}. 
Also, Table~\ref{tab:results}, shows the EER values of the same-pose experiments in the second and fourth sub-columns.

%
%
%
%
%
%

%
%

A first observation is that our SqueezeFacePoseNet model provides in general better results without the inclusion of Global Depth-wise Convolution (GDC). This is in contrast to some previous studies where GDC is reported to provide a better performance \cite{[Chen18MobileFaceNets],[Duong19MobiFace]}.
It should be mention though that the authors of our baseline networks kept the GAP layer in ResNet50ft and SENet50ft models \cite{[Cao18vggface2]}.
One possible reason of these results is that in training with VGGFace2, the face region is randomly cropped from the detected bounding box \cite{[Cao18vggface2]}, leading to images where faces are not aligned (Figure~\ref{fig:databases}c).
This may serve as an `augmentation' strategy, making counterproductive the use of GDC to learn different weights for each spatial region, since faces are not spatially aligned during training.
The use of depth-wise separable convolution (DWC) in SqueezeFacePoseNet also results in a slight decrease of performance.
This is to be expected \cite{[Howard17MobileNetv1corr]}, although it should be taken into account that adding DWC to our network reduces its model size by about 60\% (Table~\ref{tab:networks}).

Among all the networks evaluated, SENet50ft clearly stands out, specially when templates are generated with only one image (left part of Table~\ref{tab:results}), which is a much adverse case than the combination of five images (right part).
The superiority of SENet50ft over ResNet50ft for face recognition is also observed in the paper where they were presented \cite{[Cao18vggface2]}, due to the inclusion of
Squeeze-and-Excitation blocks.
Regarding SqueezeFacePoseNet, its performance is comparatively worse.
Even in that case, we believe that it obtains meritorious results, considering that it employs images of 113$\times$113 (instead of 224$\times$224), its size is $>$30 times smaller than ResNet50ft and SENet50ft, and it has $>$20 times fewer parameters.
The good results of SqueezeFacePoseNet are specially evident when using templates of five images, in whose case its EER is $<$0.55\% for any given pose, and with frontal images it is just 0.27\%.
With the lighter SqueezeFacePoseNet+DWC version, the EER of same-pose comparisons is below 1.1\%, and just 0.39\% with frontal images.

By looking at the different poses, we observe that performance decreases slightly in profile vs. profile comparisons with all networks. Even in this case, where only half of the face is visible, using templates of five images provides very good performance with any given network (EER$<$0.55\%). For the other two poses, SqueezeFacePoseNet gives a meritorious EER of 0.27/0.06\%, and an order of magnitude less is given by the baseline networks.
If templates of one image are used, our network worsens by a factor of $\sim$1.9 only w.r.t. ResNet50ft/SENet50ft.
An interesting phenomena also with all network is that the three-quarter vs. three-quarter case provides better performance than the frontal vs. frontal. 

It is also worth noting the substantial improvement observed when five images are used to generate user's templates (right part of Table~\ref{tab:results}) in comparison to using one (left part).
This points out that collecting just five images of a user is sufficient to obtain good performance across different poses with the networks employed. 
Even in a higher security scenario (e.g. FAR=0.1\%), the FRR of SqueezeFacePoseNet is below 1\% in frontal vs. frontal and three-quarter vs. three-quarter cases, and of $\sim$2\% with the lighter SqueezeFacePoseNet+DW (see Figure~\ref{fig:same_pose_results_SqueezeFacePoseNet}).
It should be considered though that the images of any user are mostly captured in different moments and they contain a very diverse variability, so the model generated when combining them is probably richer than if they were taken consecutively (e.g. from a video).
In this sense, it could be expected that the improvement would not be so high if for example we combine several shots taken consecutively, although confirming this would need extra experiments.

\subsection{Cross-Pose Comparisons}

We now carry out cross-pose verification experiments. 
Pair-wise comparisons are done between templates generated with images of different poses (Figure~\ref{fig:protocols}, right).
We follow the same protocol for 
scores generation as in Section~\ref{subsect:samepose}, resulting in the amount indicated in Table~\ref{tab:scores} (right).
Face verification results of cross-pose experiments are given in Figures~\ref{fig:inter_pose_results_SqueezeFacePoseNet} and \ref{fig:inter_pose_results_ResNet50ft_SENet50ft}.
Also, Table~\ref{tab:results} shows the EER values of the cross-pose experiments in the third and fifth sub-columns.

In a similar vein as Section~\ref{subsect:samepose}, SqueezeFacePoseNet works better in general without Global Depth-wise Convolution (GDC), and a slight performance decrease is seen when using depth-wise separable convolutions (DWC).
%
%
%
%
%
%
Also, 
SENet50ft stands out.
With SqueezeFacePoseNet, results are up to one order of magnitude worse with templates of five images, and only $\sim$1.9 times worse with
templates of one image.
Still, the EER of our network for cross-pose experiments is between 0.2-1.23\% when richer models of five images per user are employed.

Regarding the different types of poses, the worst performance is seen when there is maximum variation between the templates being compared (frontal vs. profile).
This is to be expected, given the higher variability of this combination.
Nevertheless, it should be highlighted the meritorious performance of any of the networks when templates of five images are used, with EER ranging between 0.14-1.23\% for this difficult cross-pose situation.
The best performance is always observed in the frontal vs. three-quarter case, and the three-quarter vs. profile case stands in the middle of the other two.
From these results, it can be concluded that it is not the amount of pose difference between templates that matters, but that the images appear as much frontal as possible.
In this sense, if we compare the frontal vs. frontal and frontal vs. three-quarter cases, their performance is not so different (and sometimes the frontal vs. three-quarter case is better).
In a similar vein, the frontal vs. profile is sometimes better than the profile vs. profile case. This reinforces our above observation that, in very difficult lateral poses, it is probably better to have frontal images if possible in one of the templates, rather than having all images with the same profile pose.

Similarly as Section~\ref{subsect:samepose}, using five images to generate templates is a very effective way to cope with cross-pose situations.
Its performance compared to using one image as template is significantly better (left vs. right part of Table~\ref{tab:results}), with improvements of one order of magnitude or more for any network.
In higher security situations (e.g. FAR=0.1\%), ResNet50ft and SENet50ft provide impressive FRRs below 0.5\% for any cross-pose combination, while SqueezeFacePoseNet ranks between 0.4-10\% depending on the case.

\subsection{Effect of Training Database}

We now investigate the effect of the training set in our mobile architecture (Table~\ref{tab:effects-training-db}), with all networks started from ImageNet pre-training, and trained from biometric identification as described in Section~\ref{subsec:db-protocol}.
In case that only one database is used for training,
it can be seen that better results are obtained if the model is trained on a database with more samples per user (VGGFace2), rather than on a database with more samples and more users overall but with less samples per user (MS1M).
%
%
But the biggest benefit in most cases is when the model is trained first on MS1M, and then fine-tuned on VGGFace2 (row `both').
This is in line with the results in \cite{[Cao18vggface2]}.
The biggest advantage is obtained when only one image is used to generate a user template, with improvements of up to 28\% in comparison to training on VGGFace2 only.
%
%
The effect is more diluted when five images are combined to create a user template, specially in cross-pose experiments. In this case, it is slightly better to train only on VGGFace2.
However, it is not always the case that such amount of images are always available to generate a user template, e.g. in forensics \cite{[Jain15]}.

\begin{table}[htb]
\centering

\begin{adjustbox}{max width=\textwidth}

\begin{tabular}{|c|c|ccc|ccc|c|ccc|ccc|}

\multicolumn{15}{c}{} \\ \cline{3-8} \cline{10-15}

\multicolumn{1}{c}{} &
\multicolumn{1}{c}{} &
\multicolumn{6}{|c|}{\textbf{One face image per template}} &
\multicolumn{1}{c}{} &
\multicolumn{6}{|c|}{\textbf{Five face images per template}}
\\ \cline{1-1} \cline{3-8} \cline{10-15}

\multicolumn{1}{|c|}{Training} &   & \multicolumn{3}{|c|}{\textbf{Same-Pose}} &  \multicolumn{3}{c|}{\textbf{Cross-Pose}} &  & \multicolumn{3}{|c|}{\textbf{Same-Pose}} &  \multicolumn{3}{c|}{\textbf{Cross-Pose}} \\ \cline{3-8} \cline{10-15}

\multicolumn{1}{|c|}{Data} &
 & \textbf{F-F} & \textbf{3/4-3/4} & \textbf{P-P} &  \textbf{F-3/4} & \textbf{3/4-P} & \textbf{F-P} &  & \textbf{F-F} & \textbf{3/4-3/4} & \textbf{P-P} &  \textbf{F-3/4} & \textbf{3/4-P} & \textbf{F-P}\\ \hhline{=~======~======}

MS1M &  & 16.82 & 16.23 & 20.24 &  17.45 & 21.24 & 24.19  &  & 1.17 & 2.17 & 3.25 &  1.7 & 5.24 & 7.01 \\ \cline{1-1} \cline{3-8} \cline{10-15}

VGGFace2 &  & 8.93 & 6.97 & 8.34 &  8.35 & 8.16 & 10.35  &  & \textbf{0.27} & 0.27 & 0.64 &  \textbf{0.2} & \textbf{0.55} & \textbf{1.09} \\ \cline{1-1} \cline{3-8} \cline{10-15}

both &  & \textbf{6.39} & \textbf{5.47} & \textbf{7.88} &  \textbf{6.09} & \textbf{7.02} & \textbf{8.15}  &  & \textbf{0.27} & \textbf{0.06} & \textbf{0.54} &  \textbf{0.2} & 0.88 & 1.23  \\

 & & (-28\%) & (-22\%) & (-6\%) & (-27\%) & (-14\%) & (-21\%)  &  & (-) & (-78\%) & (-16\%) & (-) & (+60\%) & (+13\%)  \\
\cline{1-1} \cline{3-8} \cline{10-15}

\multicolumn{15}{c}{} \\

\end{tabular}


\end{adjustbox}

\caption{Effect of the training database in SqueezeFacePoseNet (EER \%). F=Frontal View. 3/4= Three-Quarter. P=Profile. The best result of each column is marked in bold. Performance variation of the `both' w.r.t. the `VGGFace2' row is given in brackets.}
\label{tab:effects-training-db}

\end{table}

\section{Conclusion}

We are interested in the development of a lightweight deep network architecture capable of providing accurate cross-pose face recognition under the restrictions of mobile architectures.
For this purpose, we have adapted a very light model of only 4.41MB \cite{[Iandola16SqueezeNet]} to operate with small face images of 113$\times$113 pixels.
Training is done using the large-scale MS-Celeb-1M \cite{[Guo16_MSCeleb1M]} and VGGFace2 \cite{[Cao18vggface2]} datasets.
VGGFace2 (3.31M images, 9.1K identities) is a dataset with a rich variation of imaging conditions. Being a large-scale database, it is designed to have a larger number of images per user as well (364 on average) in comparison to other databases.
MS-Celeb-1M contains a larger number of images (3.16M in our experiments), but a larger number of identities as well (35K), so its number of images per identity is smaller.
Following recommendations \cite{[Cao18vggface2]},
we combine a large database (MS-Celeb-1M) and a database with more intra-class diversity (VGGFace2) to train the recognition network.
This has shown to provide increased performance in comparison to using only one of them (Table~\ref{tab:effects-training-db}).

To achieve further reductions in model size, we test the replacement of standard convolutions with depth-wise separable convolutions \cite{[Howard17MobileNetv1corr]}, leading to a network of just 2.5MB.
We also test Global Depth-wise Convolution (GDC) in substitution of the standard Global Average Pooling (GAP), 
since some works report that it provides better face recognition performance \cite{[Chen18MobileFaceNets],[Duong19MobiFace]}.
The employed architecture is bench-marked against two state-of-the-art architectures \cite{[Cao18vggface2]} with a size $>$30 times bigger and $>$20 times more parameters (Table~\ref{tab:networks}).
We evaluate two verification scenarios, consisting of using a different number of face images to generate a user template. In one case, a template consists of a combination of five face images with the same pose, following the evaluation protocol of \cite{[Cao18vggface2]}.
In the second case, we consider the much more difficult case of employing only one image to generate a user template.
Different pose combinations between enrolment and query templates are tested (Figure~\ref{fig:protocols}).
%
%

Obviously, the use of five face images to create a user template provides a much more better performance, with improvements of up to two orders of magnitude in some cases.
%
%
Also, in our experiments, we have not observed better performance by using Global Depth-wise Convolution, but the opposite. We speculate that this may be because training images of the VGGFace2 database are obtained by randomly cropping the face bounding box, so faces are not spatially aligned (Figure~\ref{fig:databases}c). In this sense, trying to learn different weights for each spatial region may be counterproductive.
In addition, as expected \cite{[Howard17MobileNetv1corr]}, the use of depth-wise separable convolution results in a slight decrease of performance.

Even if our light architecture does not outperform the state-of-the-art networks, it obtains meritorious results even under severe pose variations between enrolment and query templates.
For example, the comparison of frontal vs. profile images gives an EER of 1.23\%.
Also, the comparison of profile vs. profile images gives an EER of 0.54\%, even if just half of the face is visible in this case.
These results are with a template of five face images, which is revealed as a very effective way to improve cross-pose recognition performance.
With only one face image per template, the performance of our network goes up to 8.15/7.88\% respectively in the two mentioned cases.
In less extreme cases of pose variability, performance of our network is even better, for example: 0.88\% (three-quarter vs. profile view), 0.2\% (frontal vs. three-quarter), or 0.27\% (frontal vs. frontal).

A number of combinations to create enrolment and query templates would be of interest, which will be the source of future work.
For example, if video is available, a collection of frames could be combined for user template generation, probably selecting those with near to frontal pose as well.
How many images per template are necessary to obtain accurate performance is also worth to study.
In some scenarios like forensics \cite{[Jain15]}, query data may consist of only one image with an arbitrary pose, but several images per suspect may be available in the enrolment database. Therefore, one-query vs. multiple-enrolment images is also of interest to evaluate.
Also, in our protocol, a template is generated using only images of the same pose. Combining images of multiple poses in the same template could be a way to create a richer user model, further improving performance.

To improve the performance of our mobile model, we are also looking into
the use of residual connections \cite{[He16]} and pre-activation of convolutional layers inside residual blocks \cite{[He16a]}.
%
%
Giving the current context where face engines are forced to work with images of people wearing masks, we are also evaluating the accuracy of our model when using partial images containing only the ocular regions \cite{[Alonso16]}.


\section*{Acknowledgment}

This work was partly done while F. A.-F. was a visiting researcher at Facephi Biometria, funded by the Sweden's Innovation Agency (Vinnova) under the staff exchange and AI program.
Authors F. A.-F., K. H.-D. and J. B. also thank the Swedish Research Council for funding their research.
Part of the computations were enabled by resources provided by the Swedish National Infrastructure for Computing (SNIC) at NSC Linköping.
We also gratefully acknowledge the support of NVIDIA with
the donation of a Titan V GPU used for this research.

%
%
%
 \bibliographystyle{splncs04}


\begin{thebibliography}{10}
\providecommand{\url}[1]{\texttt{#1}}
\providecommand{\urlprefix}{URL }
\providecommand{\doi}[1]{https://doi.org/#1}

\bibitem{[Akhtar18-QAbiometrics]}
{Akhtar}, Z., {Hadid}, A., {Nixon}, M.S., {Tistarelli}, M., {Dugelay}, J.,
  {Marcel}, S.: Biometrics: In search of identity and security. IEEE
  MultiMedia  \textbf{25}(3),  22--35 (2018)

\bibitem{[Alonso16]}
Alonso-Fernandez, F., Bigun, J.: A survey on periocular biometrics research.
  Pattern Recognition Letters  \textbf{82},  92--105 (2016)

\bibitem{[Cao18vggface2]}
{Cao}, Q., {Shen}, L., {Xie}, W., {Parkhi}, O.M., {Zisserman}, A.: Vggface2: A
  dataset for recognising faces across pose and age. In: 13th IEEE
  International Conference on Automatic Face and Gesture Recognition (FG 2018).
  pp. 67--74 (2018)

\bibitem{[Chen18MobileFaceNets]}
Chen, S., Liu, Y., Gao, X., Han, Z.: Mobilefacenets: Efficient cnns for
  accurate real-time face verification on mobile devices. CoRR
  \textbf{abs/1804.07573} (2018), \url{http://arxiv.org/abs/1804.07573}

\bibitem{[Deng19RetinaFace]}
Deng, J., Guo, J., Zhou, Y., Yu, J., Kotsia, I., Zafeiriou, S.: Retinaface:
  Single-stage dense face localisation in the wild. CoRR
  \textbf{abs/1905.00641} (2019), \url{http://arxiv.org/abs/1905.00641}

\bibitem{[Duong19MobiFace]}
Duong, C.N., Quach, K.G., Jalata, I.K., Le, N., Luu, K.: Mobiface: A
  lightweight deep learning face recognition on mobile devices. In: Proc BTAS (Sep 2019)

\bibitem{[Guo16_MSCeleb1M]}
Guo, Y., Zhang, L., Hu, Y., He, X., Gao, J.: Ms-celeb-1m: A dataset and
  benchmark for large-scale face recognition. In: Proc ECCV. pp. 87--102. Springer
  (2016)

\bibitem{[He16]}
He, K., Zhang, X., Ren, S., Sun, J.: Deep residual learning for image
  recognition. In: Proc CVPR. pp. 770--778 (June 2016)

\bibitem{[He16a]}
He, K., Zhang, X., Ren, S., Sun, J.: Identity mappings in deep residual
  networks. In Proc ECCV. pp. 630--645. Springer International Publishing, Cham
  (2016)

\bibitem{[Howard17MobileNetv1corr]}
Howard, A.G., Zhu, M., Chen, B., Kalenichenko, D., Wang, W., Weyand, T.,
  Andreetto, M., Adam, H.: Mobilenets: Efficient convolutional neural networks
  for mobile vision applications. CoRR  \textbf{abs/1704.04861} (2017),
  \url{http://arxiv.org/abs/1704.04861}

\bibitem{[Hu18]}
Hu, J., Shen, L., Sun, G.: Squeeze-and-excitation networks. In: IEEE Conference
  on Computer Vision and Pattern Recognition, CVPR (2018)

\bibitem{[Iandola16SqueezeNet]}
Iandola, F.N., Moskewicz, M.W., Ashraf, K., Han, S., Dally, W.J., Keutzer, K.:
  Squeezenet: Alexnet-level accuracy with 50x fewer parameters and
  {\textless}1mb model size. CoRR  \textbf{abs/1602.07360} (2016),
  \url{http://arxiv.org/abs/1602.07360}

\bibitem{[Jain15]}
Jain, A.K., Ross, A.: Bridging the gap: from biometrics to forensics. Phil.
  Trans. R. Soc.  \textbf{370} (2015)

\bibitem{[Jain16]}
Jain, A., Nandakumar, K., Ross, A.: 50 years of biometric research:
  Accomplishments, challenges, and opportunities. Pattern Recogn. Lett.
  \textbf{79},  80--105 (Aug 2016)

\bibitem{[Kornblith19imagenet_transfer_better]}
{Kornblith}, S., {Shlens}, J., {Le}, Q.V.: Do better imagenet models transfer
  better? In: Proc CVPR. pp. 2656--2666 (2019)

\bibitem{[Martinez19ShuffleFaceNet]}
{Martinez-Díaz}, Y., {Luevano}, L.S., {Mendez-Vazquez}, H., {Nicolas-Diaz},
  M., {Chang}, L., {Gonzalez-Mendoza}, M.: Shufflefacenet: A lightweight face
  architecture for efficient and highly-accurate face recognition. In: Proc ICCVW. pp.
  2721--2728 (2019)

\bibitem{[Sandler18mobilenetv2]}
{Sandler}, M., {Howard}, A., {Zhu}, M., {Zhmoginov}, A., {Chen}, L.:
  Mobilenetv2: Inverted residuals and linear bottlenecks. In: Proc CVPR. pp. 4510--4520
  (2018)

\bibitem{[Sundararajan18-DLbiometrics]}
Sundararajan, K., Woodard, D.L.: Deep learning for biometrics: A survey. ACM
  Comput. Surv.  \textbf{51}(3) (May 2018)

\bibitem{[Wu18lightCNN]}
{Wu}, X., {He}, R., {Sun}, Z., {Tan}, T.: A light cnn for deep face
  representation with noisy labels. IEEE TIFS  \textbf{13}(11),  2884--2896 (2018)

\bibitem{[Zhang09SeesawNet]}
Zhang, J.: Seesaw-net: Convolution neural network with uneven group
  convolution. CoRR  \textbf{abs/1905.03672} (2019),
  \url{http://arxiv.org/abs/1905.03672}

\bibitem{[zhang19seesawfacenets]}
Zhang, J.: Seesawfacenets: sparse and robust face verification model for mobile
  platform. CoRR  \textbf{abs/1908.09124} (2019),
  \url{https://arxiv.org/abs/1908.09124}

\bibitem{[Zhang18ShuffleNet]}
{Zhang}, X., {Zhou}, X., {Lin}, M., {Sun}, J.: Shufflenet: An extremely
  efficient convolutional neural network for mobile devices. In: Proc CVPR. pp. 6848--6856 (2018)

\end{thebibliography}

\end{document}